# TA'KEED: THE FIRST GENERATIVE FACT-CHECKING SYSTEM FOR ARABIC CLAIMS

Saud Althabiti[1,2,3], Mohammad Ammar Alsalka[1,4], and Eric Atwell[1,5]

[1] School of Computing, University of Leeds, Leeds, United Kingdom
[2] Faculty of Computing and Information Technology, King Abdulaziz University, Jeddah, Saudi Arabia
[3] scssal@leeds.ac.uk , salthabiti@kau.edu.sa
[4] M.A.Alsalka@leeds.ac.uk
[5] E.S.Atwell@leeds.ac.uk

## ABSTRACT

*This paper introduces Ta'keed, an explainable Arabic automatic fact-checking system. While existing research often focuses on classifying claims as "True" or "False," there is a limited exploration of generating explanations for claim credibility, particularly in Arabic. Ta'keed addresses this gap by assessing claim truthfulness based on retrieved snippets, utilizing two main components: information retrieval and LLM-based claim verification. We compiled the ArFactEx, a testing gold-labelled dataset with manually justified references, to evaluate the system. The initial model achieved a promising F1 score of 0.72 in the classification task. Meanwhile, the system's generated explanations are compared with gold-standard explanations syntactically and semantically. The study recommends evaluating using semantic similarities, resulting in an average cosine similarity score of 0.76. Additionally, we explored the impact of varying snippet quantities on claim classification accuracy, revealing a potential correlation, with the model using the top seven hits outperforming others with an F1 score of 0.77.*

## KEYWORDS

*Ta'keed, Generative Fact-checker, Arabic Fact-Checking, Claims Verification, LLMs-based Fact-Checking, ArFactEx*

## 1. INTRODUCTION

Misinformation is false information that could be purposely shared on platforms like Twitter [1]. Users usually create it to influence what others think for political, economic, or any other reasons [2]–[4]. It can be distributed by not only individual users but also organizations and governments to negate competitors or advertise their interests [4]–[6]. Misinformation can result in significant harm, provoking confusion, causing conflict among various groups, and initiating violence [7], [8]. Therefore, it is important to remain alert and critical of the news we receive from social media and other online sources. This highlights the necessity of developing automatic fact-checking systems to ensure accuracy and reliability in the information we encounter.

Arabic Fact-checking websites, such as Fatabyyano and Misbar, verify and support claims through manual examinations and justifications to users to ensure the authenticity of some information available to the public. In contrast, most automatic Arabic fact-checking systems are developed for classification tasks, and there is a noticeable lack of studies investigating the process of providing justifications while fact-checking.

Accordingly, we aim to build an interactive tool to help users identify misinformation with clear justification. Therefore, this paper presents the Ta'keed automatic fact-checking system, which uses an LLM-based model to classify and explain a given tweet based on evidence retrieved from Google results. Additionally, this study aims to address three primary research questions related to the effectiveness and performance of the proposed fact-checking system. Firstly, it investigates the suitability of relying on snippets obtained from Google as evidential support for a given claim. Secondly, the experiments seek to determine the optimal number of snippets that improve reliability. Lastly, the study assesses the proposed system's classification accuracy through F1 scores and evaluates the proximity of the generated justifications to authentic explanations using metrics such as ROUGE and similarity measures.

The subsequent section presents the related works. Section three describes the methodology, containing system architecture, Ta'keed's interface, and system testing datasets, including ArFactEx—a new set of labelled Arabic claims with gold explanations. The experimental results are discussed in the fourth section, including the evaluation of both classification and justification generation tasks. The last section concludes this work and recommends future work.

## 2. RELATED WORK

FullFact and PolitiFact are English fact-checking websites that manually assess the accuracy of claims, with FullFact countering harm caused by false information. At the same time, PolitiFact employs a Truth-O-Meter rating system. Similarly, MISBAR and Fatabyyano are Arabic fact-checking platforms that promote reality. They serve as a leading manual source in the Arab world to distinguish between truth and falsehood by combating rumours and fake news online.

On the contrary, concerning automatic fact-checking systems, considerable studies have focused on binary or multi-class classification tasks, such as predicting a binary verdict from Arabic text [9]–[17]. However, there are few studies that explored the development of explainable systems, as observed in works like [18]. They expanded the LIAR dataset [19] by incorporating human justifications from fact-checking articles to verify claims. In contrast, [20] introduced a new dataset with journalist-crafted explanations for public health claims using extractive and abstractive summarization. Moreover, [21] utilized FEVER [22] and GPT-3 for summaries, creating the e-FEVER dataset. Lastly, [23] introduced the FactEx dataset with gold explanations and compared them with generated textual verdicts by Seq2Seq models. Nonetheless, these investigations were conducted exclusively in the English language.

In terms of similar applications, Tanbih [24], for instance, is a news aggregator that reviews articles and media sources in Arabic and English. It provides services like measuring news accuracy with propaganda levels and evaluating political bias. Additionally, Tahaqqaq [25] is a publicly available online Arabic system dedicated to assisting users in validating claims made on Twitter. In addition, it has other functionalities, including check-worthy claims identification, users' trustworthiness estimating in terms of propagating misinformation, and authoritative accounts finding. The system is integrated with the AraFacts [26] database of claims, which is periodically crawled from five Arabic Fact-checking organizations to make sure they provide support for the most recent verified claims.

Table 1. Comparative summary of the described Arabic fact-checking applications

| FCA | Method | Information Retrieval Time | Information Retrieval Sources | Classification | Explanations |
|---|---|---|---|---|---|
| Fatabyyano | Manual | - | - | ✓ | ✓ |
| Misbar | Manual | - | - | ✓ | ✓ |
| Tanbih | Auto | Continuous | Various | ✓ | - |
| Tahaqqaq | Auto | Periodic | 5 Sources | ✓ | - |
| **Ta'keed** | **Auto** | **Up to date** | **Various** | ✓ | ✓ |

On the other hand, our proposed system verifies claims based on the retrieved information from multiple sources using Google, compared with Tahaqqaq from the AraFacts dataset only. Additionally, all functionalities in Tahaqqaq tool are based on classification algorithms while we utilized an LLM-based model to not only classify a particular tweet, but also to provide an explanation for users to justify a given claim. To the best of our knowledge, Ta'keed is the only Arabic system that explains and classifies a given social media claim using an LLM-based method. Table 1 provides a comparative summary of the Arabic fact-checking applications (FCA) described in this section with the used method, the information retrieval timing and sources, and whether the system provides classification and explanations to the user.

## 3. METHODOLOGY

This section presents the architecture of the proposed system, which consists of two parts. Then, it provides a detailed description of the used datasets for testing purposes. The methodology also comprises five distinct classification experiments involving varied snippet quantities to evaluate their impact on tweet classification accuracy. We refer to each experiment as Tk#S; for instance, supporting Ta'keed with the top three snippets is called Tk3S.

### 3.1. System Architecture

The Ta'keed system mainly contains two main parts: information retrieval and claim verification with explanation.

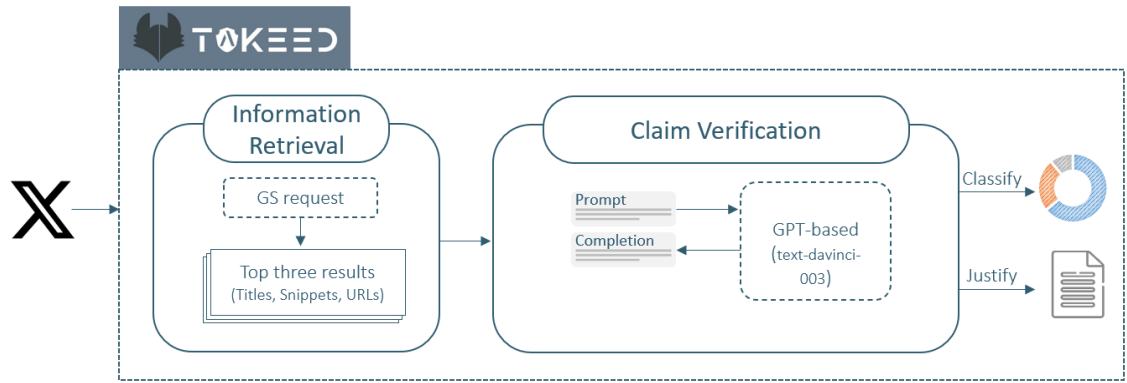

Figure 1  Ta'keed's pipeline

**Preprocessing and information retrieval:** In this part, tweets requested as queries $Q$ often contain embedded URLs; conducting a Google search $GS$ for a specific tweet yields no results, requiring the removal of irrelevant data like URLs, usernames, and hashtags as a preprocessing step. After that, we gather extra data by automatically searching each tweet on Google using requests-HTML. Each retrieved result includes the website's title, its source link, and a snippet—a brief description from Google's search results about the website's content. The pipeline demonstrated in Figure 1 illustrates the system architecture of Ta'keed.

**Claim verification:** The second part involves verifying a claim using an LLM-based model. GPT-3-based models, for example, are capable of performing various language tasks like translation, question answering, completion, and summarization [27]. We integrated the "text-davinci-003" model from OpenAI's API and set the "temperature" parameter to 7 to increase text randomness [28], as we aim to generate explanations. According to studies [21], GPT-3-based models can be effective with minimal modifications, whether fine-tuned or not. Accordingly, after conducting prompt engineering, we used the model as part of the Ta'keed system to see if it could provide sound explanations when providing the top three snippets from Google results as an initial experiment.

Algorithm 1 exhibits a role-based prompt used to instruct this model. The model is firstly prompted to perform fact-checking with a '*system*' role as the first content of the messages list *M* along with the appended instruction *I*. The second message is the Tweet *T* provided by the '*user*' is also appended to *M*. Lastly, the model with the role of '*assistant*' is to receive the snippets $S_i$ from Google results, serving as supportive information to help the model make the final classification *Cl* and explanation *tEx* returned from the completion.

---

**Algorithm 1: Prompt used to instruct the second part claim verification and explanations**

**Input:** $T_i, S_{ij}$  where $S_{ij} \in GS(Q_i)$ & $Q_i$ = pre-processed $(T_i)$
**Output:** $Cl_{ij}, tEx_i$

```
1   I ← "Assess with 'True,' 'False,' or 'Other' each tweet based on the supportive information"
2   for i = 1 to n, where n length(ArFactEx)
3       M ← {"role": "system", "content": "You are an automatic Fact Checker acting like a
            journalist clarify and" + I}
4       M ← {"role": "user", "content": Tᵢ}

5       for j = 1 to m step 2, where m length(Sᵢ)
6           Mᵢ ← ({"role": "assistant", "content": Sᵢⱼ})
7           Clᵢⱼ ← getCompletion(Mᵢ)
8       end for
9       tExᵢ ← getCompletion(M)
10  end for
```

---

### 3.2. Ta'keed Interface

We constructed the user interface using Anvil to build an interactive tool. Anvil can be used entirely with Python to build web apps using a drag-and-drop builder.

Figure 2: shows Ta'keed's interface, consisting mainly of the input, a given tweet saying, "Power outages in Al-Majaradah reach up to 20 hours," and the returned classification and justification. We adjusted the prompt to provide an English explanation in this example.

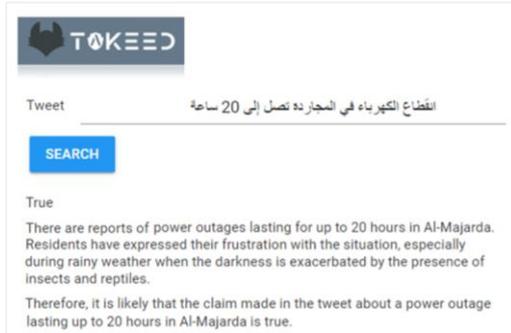

Figure 2: Ta'keed's interface with a real claim (Justification is in English in this example)

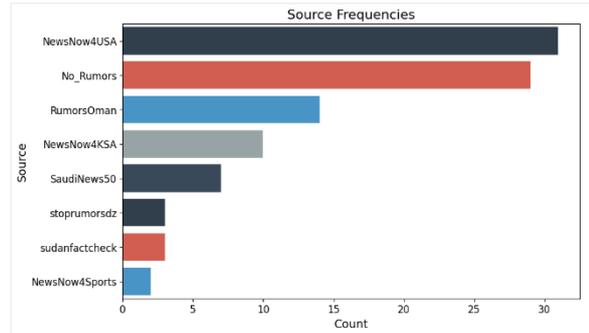

Figure 3: Collected sources distribution.

### 3.3. System Testing

#### 3.3.1. AraCOV19

As we aim to compare tweets with additional information from Google snippets [29], we employed the ArCOV19-Rumors dataset [14], [30], which comprises 138 verified COVID-19 misinformation claims sourced from reliable fact-checkers, along with more than 9,000 related tweets. These tweets were annotated to distinguish rumors from non-rumors, aiding research in detecting false information. This dataset spans from January to April 2020. We applied the first part of the system architecture (preprocessing and information retrieval) explained in section 3.1 by requesting a Google search and getting the top three results as evidence supporting the system to learn from. However, due to a problem with the labeling discussed in the results section 4, we decided to collect an additional 100 samples from trusted Arabic Twitter sources to carefully test the Ta'keed system and examine its performance in term of classification and the generated explanations.

### 3.3.2. ArFactEx Dataset

To assess the proposed system, we gathered 100 examples called ArFactEx (**Ar**abic **Fact**s with **Ex**planations) - half of which were false claims sourced from Arabic accounts like No_Rumors and stoprumorsdz, while the other half came from trusted news sources like Saudi Arabia News, labelled as either True or False news. The distribution of these sources is shown in Figure 3.

We cross-verified these labels with external news sites for accuracy. Next, we conducted manual fact-checks on each sample, providing a clear justification for the validity or inaccuracy of the claim in each tweet. We supplemented this with extended explanations from news sites and brief manual explanations based on multiple source reviews. These explanations are to provide a gold-labelled justification for each given claim or tweet. Table 2 illustrates an example containing various features: claim, source, labelling, explanation *Ex*, extended explanation *xEx*, and the sources from where the *xEx* were provided.

Table 2. A gold labelled testing example (ArFactEx instance) with its translation.

|  | **Testing example (Arabic)** | **Translation** |
|---|---|---|
| Source | هيئة مكافحة الإشاعات | No_Rumors |
| Claim (Tweet) | تقسيم شرائح استهلاك الكهرباء في السعودية الى ثلاثة أوقات في اليوم. | Dividing electricity consumption segments in Saudi Arabia into three times a day. |
| Label | False | |
| Explanation | نفت شركة هيئة تنظيم الكهرباء الخبر بشكل رسمي. | The claim was denied officially by the Electricity Regulatory Authority. |
| Extended Explanation | بينت نيوم نيوز والوئام وهيئة مكافحة الشائعات حقيقة ما يُثار حول تقسيم شرائح استهلاك الكهرباء في السعودية الى ثلاثة أوقات في اليوم. وذكرت الهيئة عبر حسابها الرسمي في «تويتر»، أن ما يتداول هو أمر غير صحيح، وسبق أن نفته رسمياً هيئة تنظيم الكهرباء. | NEOM News, Al-Weam, and the Anti-Rumor Authority revealed the truth about what is being said about dividing electricity consumption segments in Saudi Arabia into three times a day. The Authority stated, through its official Twitter account, that what is being circulated is incorrect, and it had previously been officially denied by the Electricity Regulatory Authority. |
|  | نيوم نيوز | Nuomnews |

## 4. EXPERIMENTAL RESULTS AND DISCUSSION

As an initial experiment, we utilised the ArCOV19-Rumors dataset, as each tweet is categorised as Roumor or not. We randomly selected 100 tweets with the top three related snippets collected from Google results. We first tested some of these samples using Ta'keed to determine if we could consider the collected snippets to be supportive information. We noticed that some examples were incorrectly predicted. Therefore, we examined these sample manually and found the following two reasons could be the main factors: mislabeling and insufficient retrieved information. In the first case, Table 3 presents an instance of a tweet that was classified as a rumour "False" in the published dataset, but it is incorrectly labelled according to the news articles from various news sites, while our approach could correctly classify it as "True". Secondly, more information than the snippets retrieved from Google results should be needed to make the right decision in some cases. For example, collecting more information from the news article using the URL instead of relying solely on the snippet.

Table 3. Mislabeled Example.

| Tweet | Press reports talk about Juventus striker Paulo Dybala being infected with the Corona virus | "تقارير صحفية تتحدث عن إصابة باولو ديبالا مهاجم يوفنتوس بفيروس كورونا |
|---|---|---|
| Original label | False | |
| Ta'keed's label | True | |
| Source1 (youm7.com) | Italian press reports reported that Argentine Paulo Dybala, the Italian Juventus midfielder, was infected with the Corona virus, which was announced by the World Health Organization. | أفادت تقارير صحفية إيطالية عن إصابة الأرجنتيني باولو ديبالا لاعب خط وسط يوفنتوس الإيطالي بفيروس كورونا الذي أعلنت عنه منظمة الصحة العالمية |
| Source2 (reuters.com) | Italian Football League champion Juventus said on Saturday that its Argentine striker, Paulo Dybala, was infected with the Corona virus, but did not show any symptoms. | قال يوفنتوس بطل الدوري الإيطالي لكرة القدم يوم السبت إن مهاجمه الأرجنتيني باولو ديبالا اصيب بفيروس كورونا لكن لم تظهر عليه أي أعراض ليصبح |

Therefore, we gathered additional tweets (ArFactEx testing dataset) from reliable sources and verified their validity on the Ta'keed system.

Figure 2: illustrates a successfully classified claim using the Ta'keed system with a clear explanation. While an incorrectly classified example, presented in Figure 4, shows that the system could misunderstand the Arabic claim, where it expected the word 'prohibited' to be 'there is no place'. In this instance, we adjusted the tweet but kept the claim meaning. As a result, the model could classify it correctly.

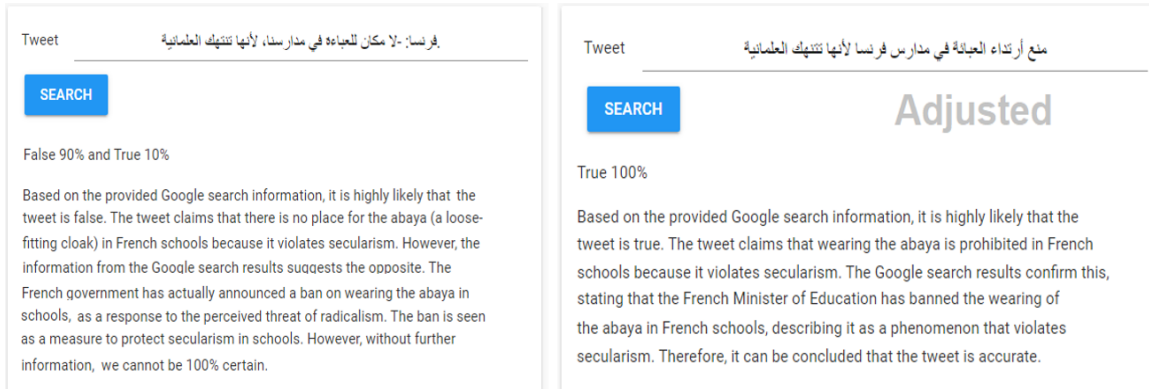

Figure 4: Example of incorrectly justified claim (The justifications are translated into English)

## 4.1. Tk3S Classification Results

The initial Tk3S model (Ta'keed based on top three snippets) showed promising results in classifying 'False' (class 0) and 'True' (class 1) instances when evaluated against the gold-labelled testing dataset (ArFactEx). It accurately identified 80% of false claims and 58% of true claims. However, since our gold label testing dataset does not have instances of class 2, it results in a 0 score for this category, as shown in the confusion matrix in Figure 5. The model predicted these instances as 'other' due to insufficient information while retrieving supportive information, making it challenging to categorize them definitively as 'False' or 'True'. Despite this, the model's performance, with an overall accuracy of 69% and an F1 score of 0.72, showcases its effectiveness in distinguishing between 'False' and 'True' claims based on the available data.

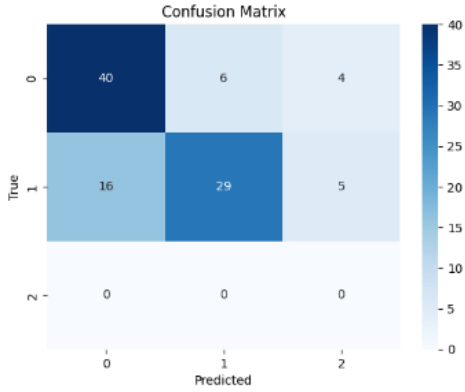
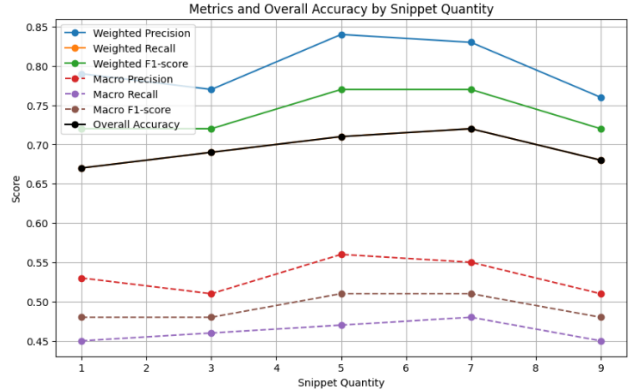

Figure 5: Tk3S classification result.

Figure 6: Scores achieved across the five models.

### 4.2. Tk3S Explanation Results

We evaluated the generated explanations against the gold references using different metrics. The results are presented in two dimensions:

**Syntactic Similarity:** The ROUGE-L-F1 scores [31] are used to compare the Ta'keed explanation *tEx* with the two sets of gold labels (explanation *Ex* and extended explanation *xEx*) shown in Figure 7. The line graph illustrates the ROUGE-L-F1 scores for each evaluated instance, providing the level of overlap and linguistic similarity between the generated text and the reference labels. The average ROUGE-L-F1 score for this comparison is only 0.15, suggesting the significance of adopting the semantic similarity.

**Semantic Similarity:** We employed cosine similarity to measure *tEx* and gold ones semantically. The results are presented in Figure 7, showing these scores for each instance in the testing dataset. The average cosine similarity score, calculated as in ( 1 ) is 0.76. This metric generally gives higher scores than the ROUGE metric, which implies a more substantial overall similarity between the generated and reference texts, suggesting this metric might be a more effective evaluation option in this case study.

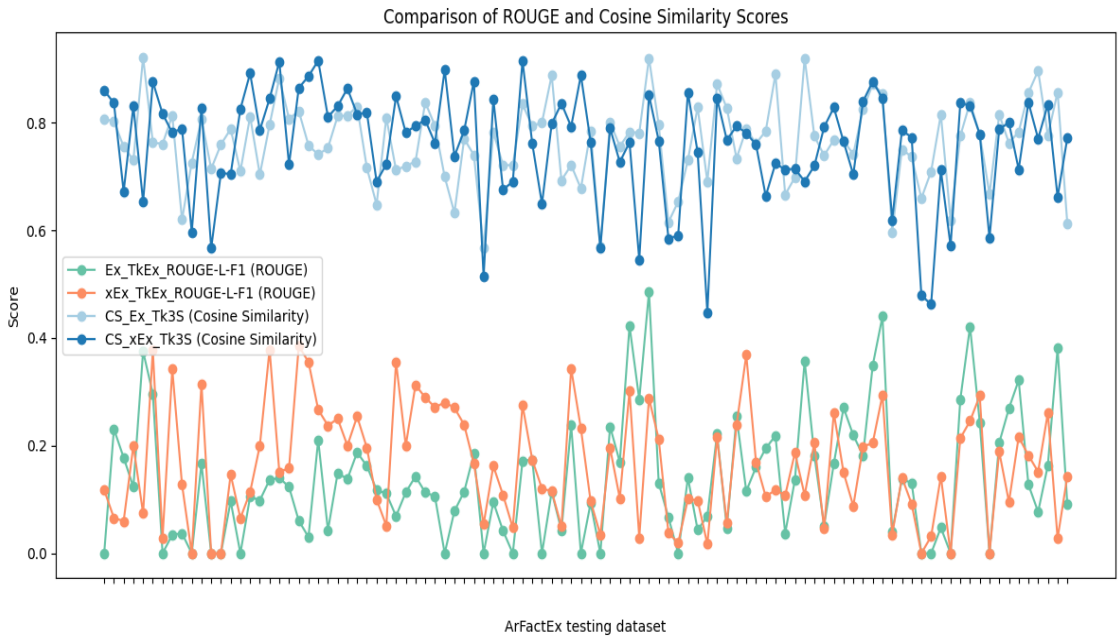

Figure 7: Syntactic and semantic similarities between *tEx* vs (*Ex* and *xEx*)

$$\bar{x} = \frac{1}{n}\sum_{i=1}^{n} \cos(tEx_i, xEx_i) \qquad (1)$$

## 4.3. Effect of Snippet Quantity on Classification Performance

This classification experiment mainly aims to assess the impact of varying snippet quantities on the accuracy of tweet classification. Each model used a different number of snippets retrieved from Google with quantities ranging from 1 to 9. The model-generated classifications were compared against the original labels from ArFactEx. Then, the performance metrics such as precision, recall, F1-score, and accuracy were computed to assess the models' classification accuracy across different snippet quantities. The results are summarized in Table 4 and Figure 8:

Table 4. Classification results across five different Ta'keed-based models

| Model | False | | | True | | | Accuracy | F1-weighted avg |
|---|---|---|---|---|---|---|---|---|
| | Precision | Recall | F1-score | Precision | Recall | F1-score | | |
| Tk1S | 0.79 | 0.68 | 0.73 | 0.79 | 0.66 | 0.72 | 0.67 | 0.72 |
| Tk3S | 0.71 | **0.80** | 0.75 | 0.83 | 0.58 | 0.68 | 0.69 | 0.72 |
| Tk5S | 0.82 | 0.72 | **0.77** | **0.85** | 0.70 | 0.77 | 0.71 | **0.77** |
| **Tk7S** | **0.83** | 0.70 | 0.76 | 0.82 | **0.74** | **0.78** | **0.72** | **0.77** |
| Tk9S | 0.77 | 0.72 | 0.74 | 0.76 | 0.64 | 0.70 | 0.68 | 0.72 |

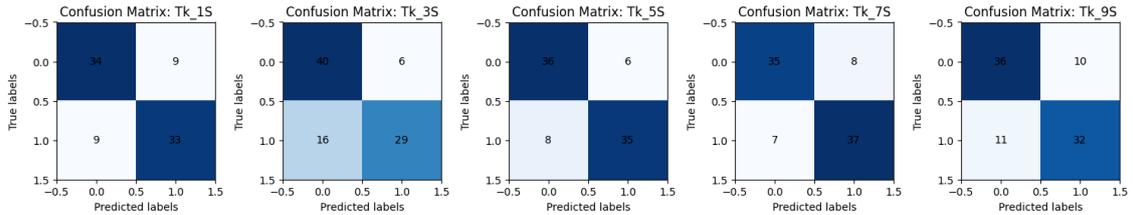

Figure 8: Confusion matrices for the five present models

The findings indicate a possible correlation between the quantity of snippets utilized and the accuracy of tweet classification. Using more snippets, specifically around 7, appears to offer a favourable balance between precision and recall, thereby enhancing the reliability of tweet classification. This approach showcases potential in leveraging external information obtained from snippets retrieved through Google to augment tweet classification models, contributing to more accurate assessments of the veracity of claims.

## 5. CONCLUSION AND FUTURE WORK

The challenge of verifying claims has intensified with the rising volume of online information. While automated fact-checking models have gained attraction for binary or multi-classification of text accuracy, there has been a notable gap in studies addressing the prediction of textual explanations, particularly in the context of the Arabic language. The primary focus of this study is developing an Arabic automatic fact-checking system, Ta'keed, which is designed to verify claims with accompanying justifications. Utilising a generative model, Ta'keed provides explanations based on information retrieved from Google search results. The system was evaluated on a sample of a previously published dataset called ArCOV19-Rumors. However, due to identified mislabelling, additional data collection was needed to investigate system performance. To assess the system, we manually assembled a gold-labelled testing dataset named ArFactEx, with justified references. The initial model demonstrated promise with an F1 score of 0.72 in the classification task. System explanations were compared with gold-standard explanations both syntactically using ROUGE and semantically. We recommend prioritising semantic evaluations, yielding an average cosine similarity score of 0.76.

Further exploration included analysing the impact of varying snippet quantities on tweet classification accuracy, revealing a potential correlation. Notably, the model's top-seven hits approach (Tk7S) outperformed others, achieving an F1 score of 0.77. This comprehensive evaluation highlights the effectiveness of Ta'keed in providing reasonable classification results with meaningful explanations for claim verification.

In future work, we intend to expand the application of this method to diverse datasets containing various languages beyond Arabic. Moreover, in line with our suggestion to improve the retrieval part, we aim to explore not just Google snippets but also consider additional details in news articles retrieved from webpages using their respective URLs.

## ACKNOWLEDGEMENTS

We sincerely thank the Ministry of Education in Saudi Arabia, King Abdulaziz University, and the University of Leeds for their support.